%% file: accv2020submission.tex
\documentclass[runningheads]{llncs}
\usepackage{graphicx}
\usepackage{amsmath,amssymb} 
\usepackage{color}
\usepackage[table,xcdraw]{xcolor}
\usepackage[pagebackref=true,breaklinks=true,letterpaper=true,colorlinks,bookmarks=false,citecolor=blue,linkcolor=blue]{hyperref}
\usepackage{caption} 
\begin{document}
\pagestyle{headings}
\mainmatter


\title{Unsupervised Discovery of Disentangled Manifolds in GANs} 


\author{Yu-Ding Lu, Hsin-Ying Lee, Hung-Yu Tseng, Ming-Hsuan Yang}
\institute{University of California at Merced}
%
%

\maketitle

\begin{figure}[h]
    
    \centering
    \includegraphics[width=12cm]{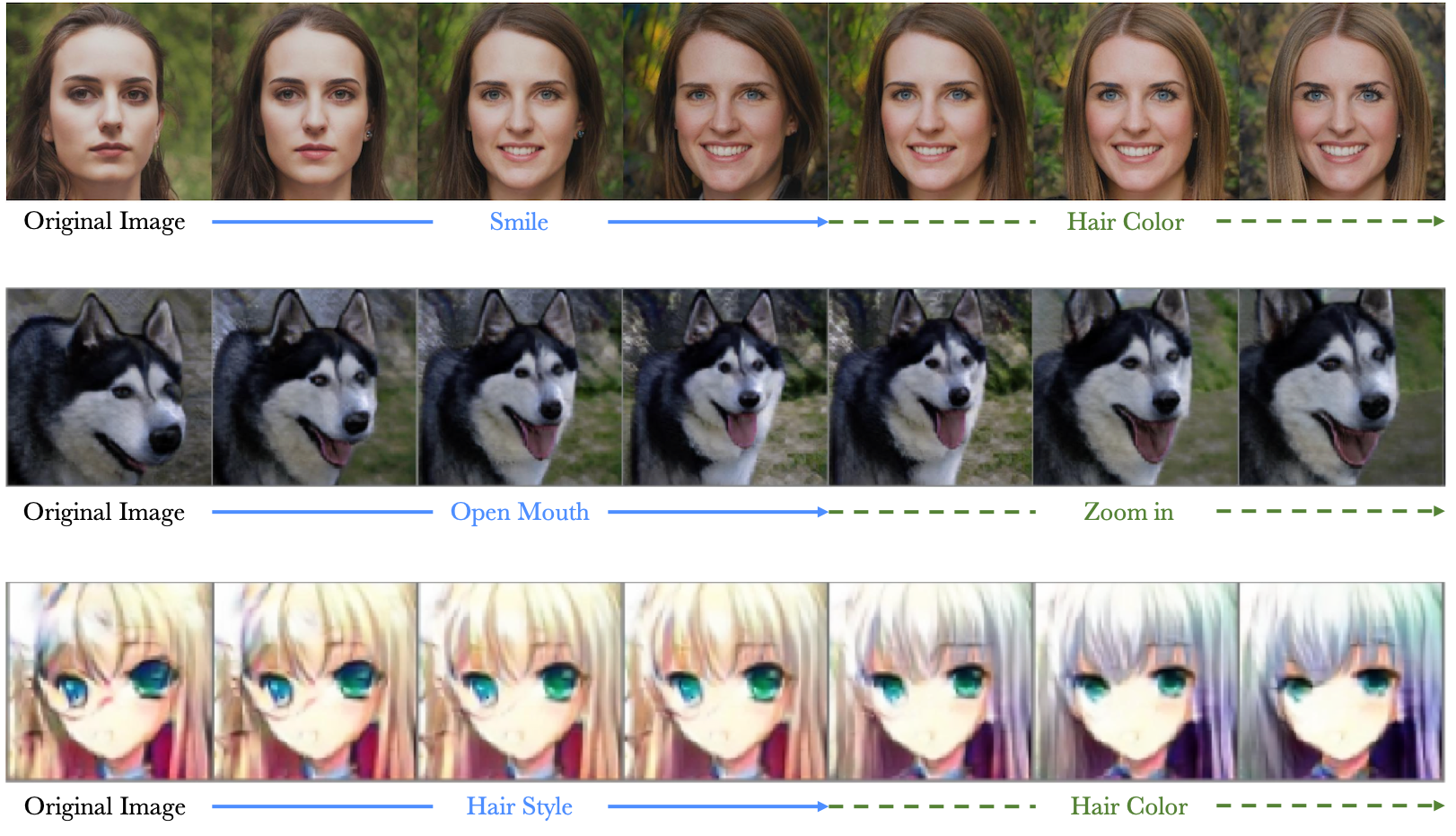}
    \caption{\textbf{Interpretable latent space discovery.} 
    The proposed method explores the interpretable directions in the latent space of pretrained models in an unsupervised manner.
    With the mined interpretable directions we can manipulate images by changing different attributes smoothly.
    }
    \label{fig:teaser}
\end{figure}

\begin{abstract}
As recent generative models can generate photo-realistic images, people seek to understand the mechanism behind the generation process.
Interpretable generation process is beneficial to various image editing applications.
In this work, we propose a framework to discover interpretable directions in the latent space given arbitrary pre-trained generative adversarial networks.
We propose to learn the transformation from prior one-hot vectors representing different attributes to the latent space used by pre-trained models.
Furthermore, we apply a centroid loss function to improve the consistency and smoothness while traversing through different directions.
We demonstrate the efficacy of the proposed framework on a wide range of datasets.
The discovered direction vectors are shown to be visually corresponding to various distinct attributes and thus enable attribute editing.

%
%

\end{abstract}

\input{1_intro}

\input{2_related}

\input{3_method}

\input{4_experiment}

\input{5_conclution}

\bibliographystyle{splncs}
\bibliography{egbib}

\end{document}

%% file: 1_intro.tex
\section{Introduction}

Image generation has achieved great success with the help of Generative Adversarial Networks (GANs).
We can generate photo-realistic images in both unconditional~\cite{karras2019style} and conditional~\cite{brock2018large} manner.
Recently, interpretability of the latent space where generators sample noise vectors has getting attention.
A controllable latent space can enable various editing applications like object insertion, object removal, and the attribute altering.

To enable such control over the latent space, disentanglement of each dimension in the latent space is desirable.
The core concept of disentanglement is to associate different dimensions in the latent space with distinct and informative properties of the target domain.
We can thus manipulate images to change a specific attribute by controlling corresponding dimensions without altering other attributes.

A variety of methods for achieving disentanglement have been proposed.
A branch of work aims to enforce disentanglement during training by maximizing the mutual information between a small subset of latent variables and observation~\cite{chen2016infogan}, applying mixing regularization using two latent codes~\cite{karras2020analyzing}, and restricting the influence of each dimension to specific parts of synthesized image by feeding multiple noise codes through separate fully-connected layers.
The other branch of work attempts to explore the latent space of a pre-trained GAN model by identifying interpretable directions~\cite{voynov2020unsupervised}.

In this work, we propose a novel framework to exploit the latent space of pre-trained GAN models in an unsupervised manner.
We sample from the latent space two latent vectors from different dimensions and with different lengths.
Given an original image generated by the original latent code, we perform first translation by shifting the original latent code with the first latent vector.
The generated image is then translated to another image by shifting the vector again with the second latent vector.
Our model is asked to predict the direction and the scale of these two latent codes given the original image and two shifted images.
Moreover, we use centroid loss to make deformator learn the centroid features for each direction and penalize the distances between the shifted codes and their corresponding centers. 

We conduct extensive quantitative and qualitative experiments to validate the efficacy of the proposed framework.
We apply the proposed framework on the Anime Faces, ILSVRC-ImageNet and Flickr-Faces-HQ datasets using SNGAN, BigGAN and StyleGAN2 respectively.
We use Reconstructor Classification Accuracy (RCA) and Perceptual Path Length (PPL) metrics.
We perform better in diversity and consistency.
Qualitatively, we observe that our approach can smoothly change attributes from one direction to another.

We make the following contribution in this work:
First, we introduce a training strategy to generate consecutive images from one direction to the next.
Second, we apply centroid loss to keep the shifted codes in the same direction being close to the centroid of the direction in latent space.
Finally, we validate our results on different datasets and pre-trained generator. It reveals our approach is generalized in this task.

%% file: 2_related.tex
\section{Related work}

\subsection{Generative adversarial networks}
Generative adversarial networks~(GANs)~\cite{karras2019style,arjovsky2017wasserstein,goodfellow2014generative,karras2017progressive,lee2020drit++} have been widely used for the image generation task. 
The core idea is to encode the image distribution into a latent space by learning the mapping from latent representations to synthesized images via adversarial learning.
%
%
To interpret the latent space, the InfoGAN approach~\cite{chen2016infogan} maximizes the mutual information to disentangle the latent space in an unsupervised fashion.
The FineGAN~\cite{singh2018finegan} and VON~\cite{zhu2018visual} model uses a hierarchy architecture to encode different types of variation into separated latent spaces.
In this work, we explore to identify interpretable directions in the latent space of pre-trained GANs. We thus discover different directions to control the attributes in generated images.

\subsection{Interpolations in latent space}
The recent years have witnessed growing interests in developing to find the interpretable latent space for GANs that corresponds to different attributes.
StyleGAN2 disentangle style and content representations of the data \cite{karras2020analyzing}. They propose a path-length regularized and revised architecture to make GANs easier to attribute a generated image to its source. 
\cite{Jahanian*2020On} study the ability of transformation and generate realistic images by steering features in latent space. They also quantify the limitation of GANs transformations and mitigate the problem. 
\cite{Deng_2020_CVPR} embed 3D prior into adversarial training and generated faces can be precisely controlled.
\cite{Alharbi_2020_CVPR} inject multiple spatially-variable codes to change the attributes on generated images through grid-based structures
Also, \cite{voynov2020unsupervised} propose an unsupervised way to explore interpretable directions in the latent space of a pre-trained GAN model.
In our approach, we expand \cite{voynov2020unsupervised} approach by modifying the framework and introduce a centroid loss to improve the quantitative results and generate diverse images with semantic attributes.

%% file: 3_method.tex
\begin{figure}[h]
    
    \centering
    \includegraphics[width=10cm]{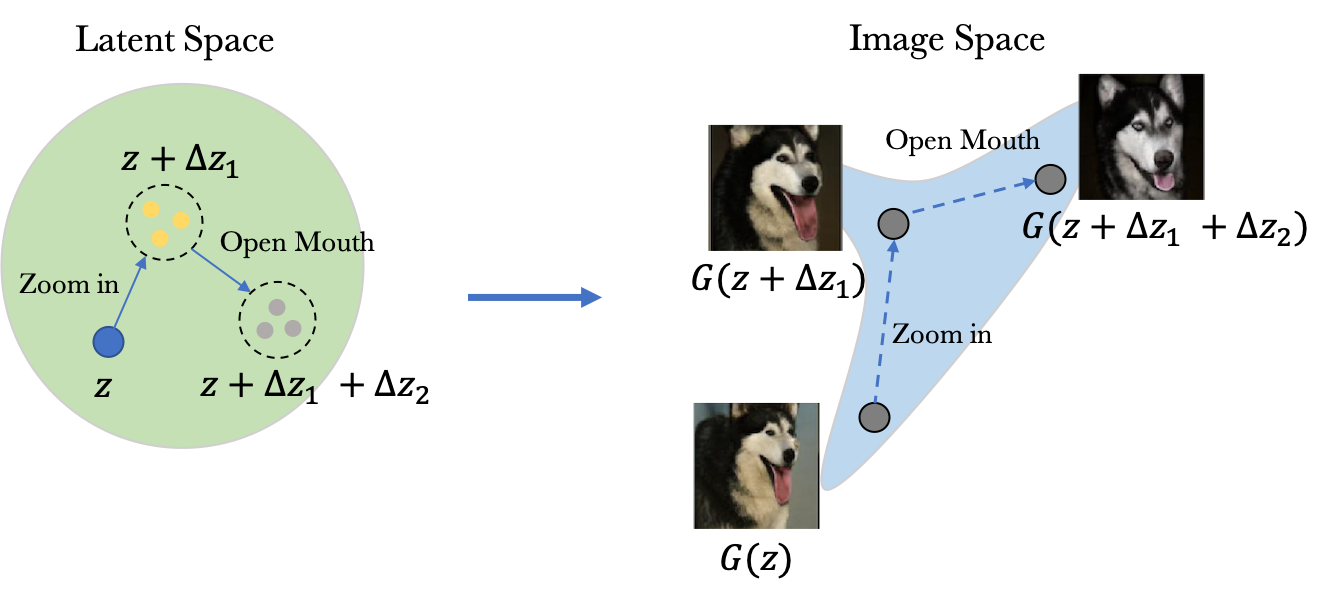}
    \caption{\textbf{Concept Illustration.} Our method aims to discover interpretable directions in latent space in an unsupervised manner.
    For example, we can transform the generated image $G(z)$ to $G(z+\Delta z_1)$ and $G(z+\Delta z_1+\Delta z_2)$ with semantic attributes. }
    \label{fig:concept}
\end{figure}

\section{Proposed method}
Given a pre-trained generator, the goal is to find the interpretable directions in the latent space for the manipulation purpose.
We present an overview of the proposed approach in Figure~\ref{fig:model}.
In addition to the pre-trained generator $G$, our design consists of a deformator $A$ and a reconstructor $R$. 
Given the desired manipulation attribute $k$ (e.g., zoom in) and the magnitude of the manipulation $\varepsilon$, the deformator $A$ aims to find the interpretable direction in the latent space learned by the pre-trained generator $G$.
After the latent representation $z$ is shifted by the deformator $A$, the pre-trained generator $G$ synthesizes the manipulated image. 
According to the original and manipulated images, the reconstructor $R$ recovers the intended manipulation attribute $\hat{k}$ and magnitude $\hat{\varepsilon}$.

\subsection{Learning how to shift}

\begin{figure}[h]
    \centering
    \includegraphics[width=12cm]{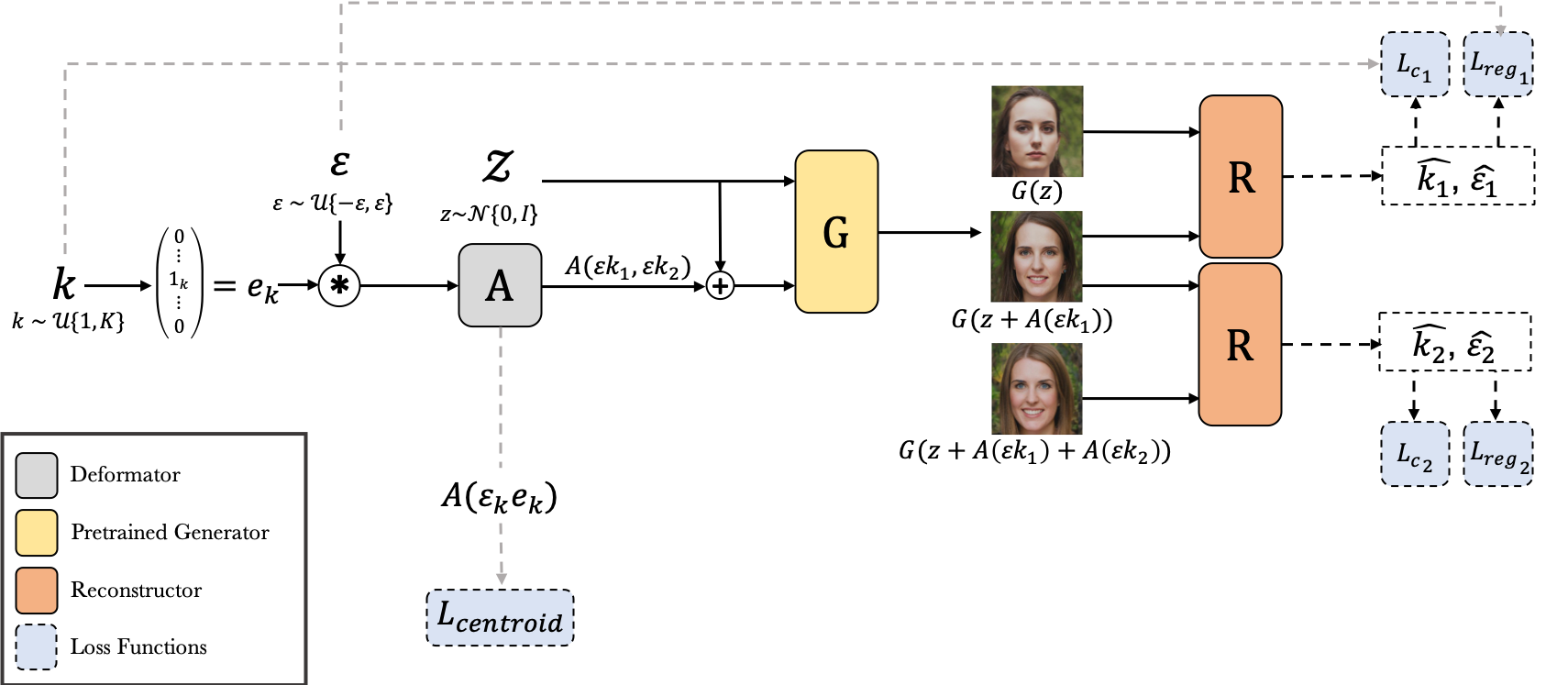}
    \caption{
    \textbf{The proposed framework.}
    The framework consists of a pre-trained generator $G$, a deformator $A$, and a reconstructor $R$.
    The deformator $A$ map the sampled $\varepsilon$ and $k$ to the discriminative shifted codes $A(\varepsilon k)$ in latent space. The reconstructor $R$ aims to reconstruct $k$ and $\varepsilon$ from the generated image and shifted image pair. Our shifting training adds one shifted codes in latent space to provide the deformator find linearly shifted codes and apply these to a generator. Also, we apply centroid loss to cluster shifted codes with the centroid of semantically meaningful direction in latent space.}
     \label{fig:model}
\end{figure}

Since the latent space learned by a pre-trained generator is usually not interpretable.
To address the issue for the image editing purpose, we employ a deformator $A$ to calculate the shifted feature (i.e., manipulation direction) in the latent space according to the user-intended editing attribute $k$ and magnitude $\varepsilon$.
%
%
Specifically, the deformator consists of three fully-connected layers to map input editing attribute $k$ and magnitude $\varepsilon$ to the shifted features.
%
The editing attribute $k$ is a real number ranging from $0$ to $128$, where $128$ represents the number of attributes we discover for the manipulation.
As shown in the left-hand side of Figure~\ref{fig:model}, we convert the real number $k$ to the one-hot vector $e_k$, and multiply by the magnitude term $\varepsilon$ as the input of the deformator $A$.
Assuming the original image $G(z)$ is generated from the latent representation $z\sim N(0, I)$, we shift $z$ with the prediction from the deformator, namely $z + A(e_k\varepsilon)$.
The generator then synthesizes the manipulated image $G(z+A(e_k\varepsilon))$.

%
%
%
To ensure the manipulated image $G(z+A(e_k\varepsilon))$ corresponds to the intended editing, we design a reconstructor $G$ to recover the input attribute $k$ and magnitude $\varepsilon$, as shown in the right-hand side of Figure~\ref{fig:model}.
Specifically, the reconstructor $R$ takes as input the original image $G(z)$ and the manipulated image $G(z+A(e_k\varepsilon))$, and predict the editing attribute $\hat{k}$ and magnitude $\hat{\varepsilon}$, which can be formulated as follows:
\begin{equation}
    \hat{k}, \hat{\varepsilon} = R(G(z), G(z+A(e_k\varepsilon))),
\end{equation}
where $e_k$ is the one-hot representation of the editing attribute $k$.
Finally, the recovered editing attribute and magnitude should be identical to the input ones.

%
%

\subsection{Training loss functions}
To extend the flexibility and diversity for searching directions in latent spaces, we apply two strategies to improve this model: linearly shifting and centroid loss. 
First, as demonstrated in the Figure \ref{fig:model}, we respectively synthesized three images $G(z)$, $G(z+A(e_{k1}\varepsilon_1))$ and $G(z+A(e_{k1}\varepsilon_1)+A(e_{k2}\varepsilon_2))$ from pre-trained GANs. 
In order to show the capacity of deformator can change the attributes linearly for shifted images, we then minimize the following loss function:

\begin{equation}
\min_{A,R} \mathbb{E}_{z,k_{1,2},e_{1,2}} L(A, R) = \min_{A,R} \mathbb{E}_{z,k_{1,2},e_{1,2}}[L_{cl}(k_{1,2}, \hat{k}_{1,2})+\lambda L_{r}(\varepsilon_{1,2}, \hat{\varepsilon}_{1,2})]
\end{equation}

The classification loss $L_{cl}$ is cross-entropy to classify the direction of input image pairs. Next, the mean absolute error $L_{r}$ is used to predict the magnitude value in shifted images. 
To ensure the shifted codes in latent space is linear and scalable, we not only calculate losses between image pairs from synthesized images $G(z)$ and shifted images $G(z+A(e_{k1}\varepsilon_1))$ but also calculate losses between $G(z+A(e_{k1}\varepsilon_1))$ and $G(z+A(e_{k2}\varepsilon_2))$.

Second, in order to make shift direction be more consistent in latent space, we present a loss function to find the centroid for shift codes. Similar to previous work \cite{wan2018generalized,7953183,wen2016discriminative}, the centroid of the shift vectors from the $k$th dircetion is defined as $c_k$ via following equation:

\begin{equation}
   c_k = \sum^M_{m=0} A(e_{k}\varepsilon)
\label{eq:centroid}
\end{equation}
The $M$ is the total number of shift codes of $k$ direction. When we train this model, the $M$ is added by the number of shift code in a batch. 
We then calculate the similarity between shift code and its centroid using the cosine similarity:

\begin{equation}
   S = cos(A(k\varepsilon), c_k)
\end{equation}  

The $S$ represents the similarity between shift code and its centroid. Thus, we can define the centroid loss as :

\begin{equation}
   L_c = \frac{1}{N}\sum_{n=1}^{N} S(A(k\varepsilon), c_k)
\end{equation}  

$N$ denotes the total number of training samples in a batch. The formulation effectively characterizes the intra-class variations. Ideally, the $c_k$ should be iteratively updated during the training steps. In each iteration, the centroids are computed by Equation \ref{eq:centroid} for the corresponding classes. 

As a result, the total loss in our approach is as the following:

\begin{equation}
   \min_{A,R} \mathbb{E}_{z,k_{1,2},e_{1,2}} L(A, R) = \min_{A,R} \mathbb{E}_{z,k_{1,2},e_{1,2}}[L_{cl}(k_{1,2}, \hat{k}_{1,2})+\lambda L_{r}(\varepsilon_{1,2}, \hat{\varepsilon}_{1,2})+\gamma L_c]
   \label{eq:total}
\end{equation} 

In all our experiment, we use $\lambda=0.5$ and $\gamma=0.25$. Also, the deformator is a fully-connected network to deform the features for shifting noise. Thus, we can jointly optimize $A$ and $R$.

%% file: 4_experiment.tex
\section{Experiment}
In this section, we conduct several experiments to evaluate the proposed approach. We qualitatively and quantitatively compare our method with baseline model on different metrics including Reconstructor Classification Accuracy(RCA) and Perceptual path length (PPL). Also, we highlight our key findings with our training strategies. Experimental results reveal that our method can produce more consistent and diverse images. More comparisons are provided in the supplementary material.  
\subsection{Evaluation metrics}
\subsubsection{Reconstructor Classification Accuracy (RCA)}
As noted by \cite{voynov2020unsupervised}, they present an evaluation metric for finding directional shift code for reconstructor $R$.
The $R$ aims to predict what direction and magnitude between shifted image and synthesized image. Thus, the RCA aims to calculate the accuracy of the prediction in the reconstructor. In essence, the high RCA score is related to the shifted images that are discriminative and easy to distinguish from each other. 
In our experiment, RCA allows us to compare our method with the baseline method\cite{voynov2020unsupervised}. 
However, the RCA score only implies that the corresponding image transformation and it cannot evaluate the performance of disentanglement. Thus, we need to use another metric to calculate how serious changes the image when we shift code in latent space. 

\subsubsection{Perceptual path length (PPL)}
The perceptual path length \cite{karras2019style} measures the difference between consecutive images when interpolating between two inputs. In our experiment, we generate consecutive images by shifting the value in the latent codes.  
The PPL denotes a less curved latent space should result in a lower value than highly curved latent space. In essence, it infers the variation between the image transformation in our approach. 
Following the PPL in \cite{karras2019style}, we use PPL to calculate the pairwise image distance between the shifted image and synthesized image by using VGG16\cite{Simonyan15} embeddings. In each dataset, we compute the expectation by taking 10,000 samples. 

\begin{figure}[h]
    
    \centering
    \includegraphics[width=12cm]{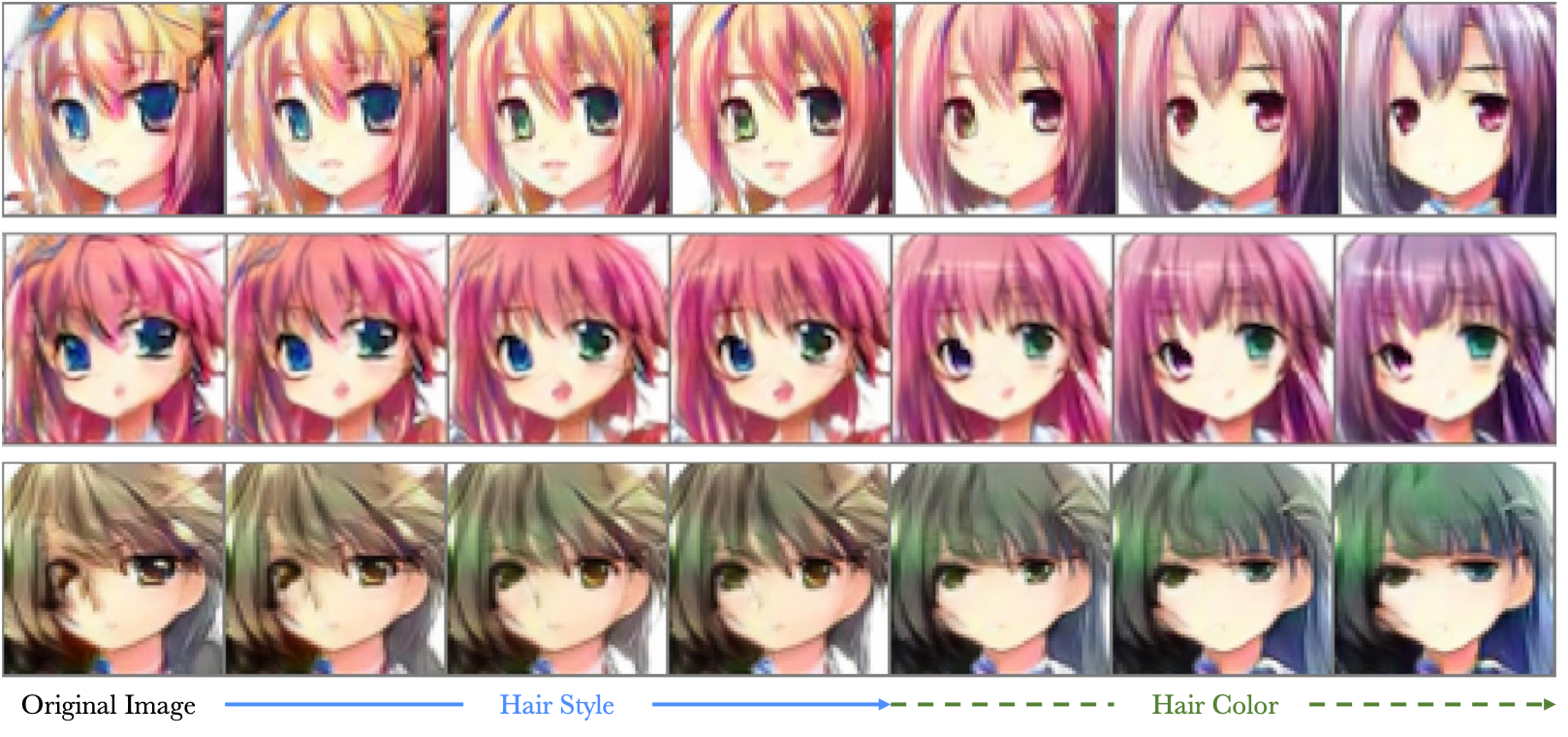}
    \caption{
    \textbf{Sampled results.}
    Examples of interpretable directions for Spectral Norm GAN on the Anime Faces dataset. 
    The synthesized images changes attributes from left to right. 
    Rows showcase the transformation of the original images along with two different attribute directions.}
    \label{fig:anime}
\end{figure}

\subsection{Implementation Details}
Our model is implemented in Pytorch. By following baseline model \cite{voynov2020unsupervised}, we set the number of directions $K$ to 128 for all experiments to explore different interpretable directions in latent spaces. 
In order to make the $A$ be flexible to fit latent space with different pre-trained generator, we use 3 fully-connected layers to train $A$. The number of hidden nodes are set to 1024.
For the reconstructor $R$, following the previous work, we use the LeNet\cite{lecun1998gradient} for Anime Faces dataset and the ResNet-18\cite{he2016deep} for Imagenet and FFHQ. We concatenate image pairs and set input channel to 6.
The distribution of training samples are sampled from normal distribution and the direction index $k$ and shift magnitude $\varepsilon$ are set to the uniform distribution $\mathcal{U}[1, K], \mathcal{U}[-6, 6]$ respectively. 
In all the experiments, we use Adam optimizer \cite{kingma2014adam} to optimize our approach with a constant learning rate 0.0001 and $2*10^5$ gradient steps.

\subsection{Comparative experiments}

\subsubsection{Anime Faces and SNGAN} this dataset\cite{jin2017towards} includes 63632 high-quality anime faces images. We used all images as our training data and reshape images to 64 $\times$ 64 to train the Spectral Normalization GAN(SNGAN)\cite{miyato2018spectral} to be pre-trained model. The SNGAN includes spectral normalization as a stabilizer in GANs. It consists of ResNet-like generator of three residual blocks. 
In our experiment, we train a SNGAN by using WGAN-GP\cite{gulrajani2017improved} to provide stable and consistent synthesized images. We then train our approach on this pre-trained generator. 
In Figure \ref{fig:anime}, it shows the qualitative examples of transformation with our method. The left side is the first synthesized image from $z$ and we fixed $k_1, k_2$ from $K$ and gradually shifted magnitude $\varepsilon_1, \varepsilon_2$ to generate shifted images from $G(z+A(k_1, \varepsilon_1)$ to $G(z+A(k_1, \varepsilon_1)+A(k_2, \varepsilon_2)$. 
The result reveals that the deformator $A$ can effectively change the attributes in synthesized images. The synthesized image first changes the style of hair and then alters the color of hair. 


\begin{figure}[th!]
    
    \centering
    \includegraphics[width=12cm]{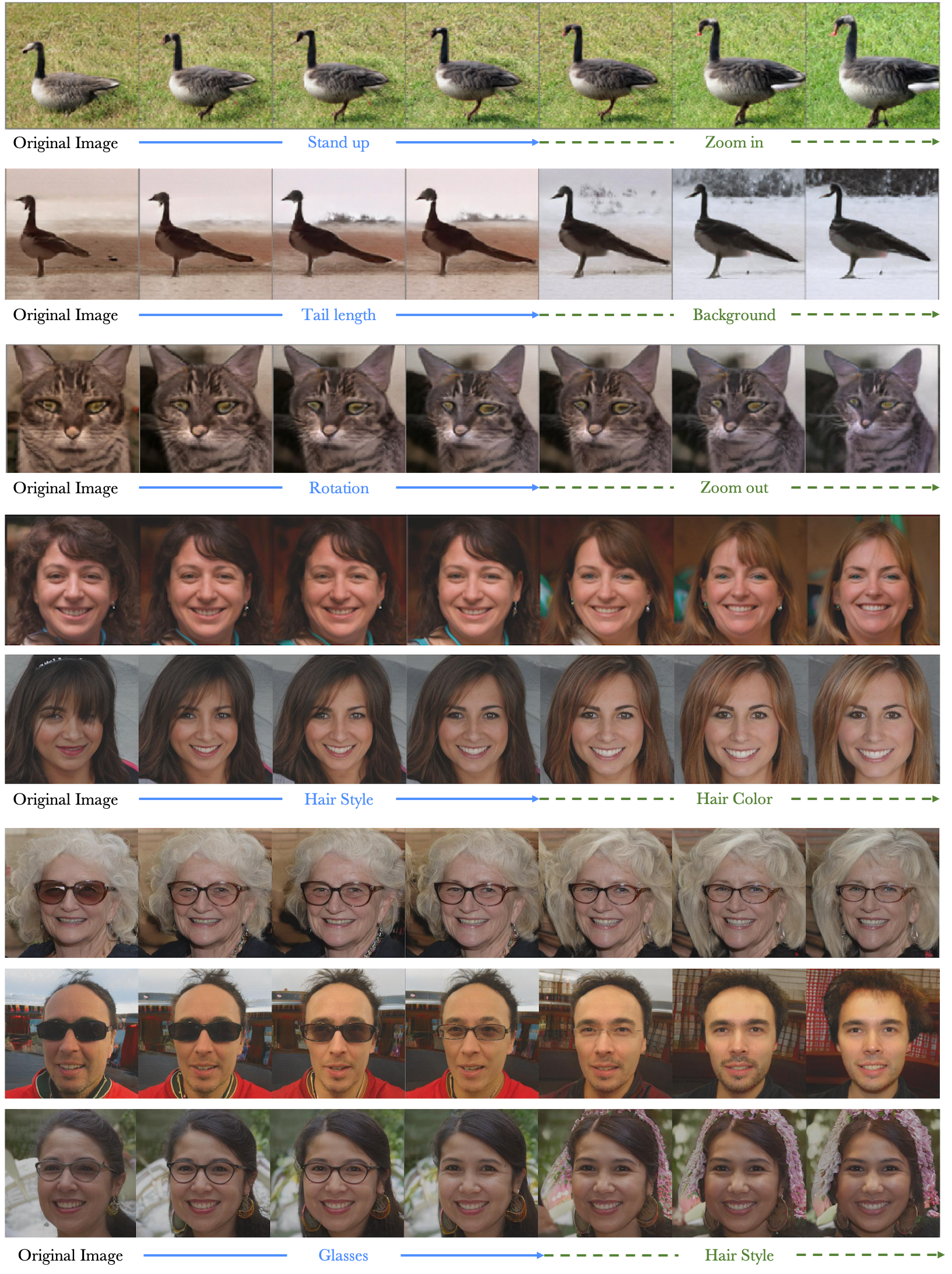}
    \caption{
    \textbf{Diversity of discovered directions.}
    The results show some examples of discovered interpretable directions on the ILSVRC and FFHQ datasets using pre-trained BigGAN and StyleGAN2, respetively. These images demonstrate the diversity of attributes discovered by the proposed method.}
    \label{fig:Biggan_FFHQ}
\end{figure}

\subsubsection{ILSVRC-ImageNet and BigGAN} We use ILSVRC-ImageNet\cite{ILSVRC15} dataset, containing large-scale images with 1000 categories to train the pre-trained BigGAN\cite{brock2018large} generator to synthesize our images at 128 $\times$ 128. BigGAN is the first large-scale generator that can generate high-fidelity quality images with "truncation trick". This architecture is able to synthesize high-quality images with conditional vectors. 
In our experiment, in order to reduce reconstructor difficulty of predicting the direction and shift magnitude, we fixed conditional vector to train deformator $A$ and reconstructor $R$. Thus, we need to train our model for different categories in this dataset. 
Figure \ref{fig:Biggan_FFHQ} shows several interesting directions in shifted codes. For example, the direction of goose image in the first row corresponding to the motion of object and zoom in the images.
These results reveal that we can synthesize images with different attributes and shift images consistently from one direction to another.

\subsubsection{Flickr-Faces-HQ and StyleGAN2} The Flickr-Faces-HQ dataset includes 70,000 high-quality face images at 1024 $\times$ 1024 resolution. This paper \cite{karras2020analyzing} presents a path length regularized generator, is called StyleGAN2, to synthesize high-quality face images and we use a pre-trained model from online.
Due to the StyleGAN2 mainly synthesized images from intermediate latent space $W$, we add our shifted codes to the $W$ in each layer of the generator.
We demonstrate generated images of discovered directions for StyleGAN2 in Figure \ref{fig:Biggan_FFHQ}. It reveals that our model can discover several attributes in latent space. For instance, the shifted codes are able to change color of hair, style of hair or wearing glasses or not.

    
    

\begin{table}[th!]

\centering
\begin{tabular}{|l|c|c|c|c|c|c|}
\hline
                               & \multicolumn{2}{c|}{\cellcolor[HTML]{FFFFFF}Anime Faces} & \multicolumn{2}{c|}{\cellcolor[HTML]{FFFFFF}ILSVRC} & \multicolumn{2}{c|}{\cellcolor[HTML]{FFFFFF}FFHQ} \\ \hline
Method                         & \textbf{ RCA$\uparrow$ }                & \textbf{ PPL$\downarrow$ }               & \textbf{ RCA$\uparrow$ }             & \textbf{ PPL$\downarrow$ }             & \textbf{ RCA$\uparrow$ }            & \textbf{ PPL$\downarrow$ }            \\ \hline
Baseline                       & 0.77                        & 179.23                     & 0.84                     & 207.35                   & 0.84                    & 412.40                  \\ \hline
Ours w/o Centroid Loss                 & 0.80                        & 175.32                     & 0.86                     & 191.34                   & 0.83                    & 408.75                  \\ \hline
Ours & 0.78                        & 99.78                      & 0.85                     & 185.23                   & 0.82                    & 402.64                  \\ \hline
\end{tabular}
\caption{\textbf{Quantitative comparison.} RCA evaluates the accuracy of reconstructor. High RCA values denote that the interpretable directions are easy to distinguish by reconstructor. The PPL measures the perceptual distance of consecutive images with interpolation in the latent space. The lower values result in a perceptually smoother transition than higher values.}
\label{tb:results}
\end{table}

\subsubsection{Quantitative comparison}
For quantitive comparison, we use RCA and PPL to evaluate the quality of shifted images. Here we perform the RCA and PPL to observe our results. The high RCA value indicates that discovered directions are discriminative and lower PPL means consecutive images are less different from each other.
In Table \ref{tb:results}, we compare our model with two algorithms 
the RCA score of our approach is competitive with the baseline model. Namely, our approach increases the flexibility to shift code from one direction to another and it did not affect the reconstructor to distinguish the difference between synthesized images.
Furthermore, our approach obtains a higher score on PPL since the centroid loss implies the shifted codes in the same direction $k$ would be close to the centroid of $k$. To be more specific, the shifted images from the same $k$ can keep more characteristic than the baseline model.

%% file: 5_conclution.tex
\section{Conclusion}
In this paper, we present a novel unsupervised method to discover the interpretable directions in the GAN latent space. 
We propose to add another shifted image pairs to generate consecutive images with different semantic attributes and apply centroid loss to cluster intra-class in latent space.
We demonstrate our approach on several datasets and the experimental results reveal that the pre-trained generator is able to synthesize samples from multiple directions in shifted codes. 
Compared with the existing method, our approach has the advantage of generating images with diverse attributes and keeping the perceptual difference less between the synthesized image and the shifted images.
In future work, we will continue to apply different strategies to effectively discover the directions of latent space in GANs.
%